%% file: main.tex
\documentclass[10pt,twocolumn,letterpaper]{article}

\usepackage[pagenumbers]{cvpr} 

\usepackage{graphicx}
\usepackage{amsmath}
\usepackage{amssymb}
\usepackage{booktabs}
\usepackage[accsupp]{axessibility}  

\usepackage[pagebackref,breaklinks,colorlinks]{hyperref}

\usepackage[capitalize]{cleveref}
\usepackage{color, colortbl}

\crefname{section}{Sec.}{Secs.}
\Crefname{section}{Section}{Sections}
\Crefname{table}{Table}{Tables}
\crefname{table}{Tab.}{Tabs.}

\newcommand{\hy}{\hat{y}}
\newcommand{\hb}{\hat{b}}

\newcommand{\noobject}{\varnothing}

\newcommand{\R}{\mathbb{R}}

\newcommand{\bc}{\boldsymbol{c}}

\newcommand{\bq}{q}

\newcommand{\bx}{\boldsymbol{x}}
\newcommand{\bv}{\boldsymbol{v}}

\newcommand{\T}{\mathcal{T}}

\newcommand{\F}{\mathcal{F}}

\def\cvprPaperID{7204} 
\def\confName{CVPR}

\begin{document}

\title{MUTR3D: A Multi-camera Tracking Framework via 3D-to-2D Queries}

\author{
  Tianyuan Zhang \\ 
  Carnegie Mellon University \\
  \texttt{tianyuaz@andrew.cmu.edu} \\
  \and 
  Xuanyao Chen \\
  Fudan University \\
  \texttt{xuanyaochen19@fudan.edu.cn} \\
  \and
  Yue Wang\\
  Massachusetts Institute of Technology \\
  \texttt{yuewang@csail.mit.edu} \\
  \and
  Yilun Wang \\ 
  Li Auto \\ 
  \texttt{yilunw@cs.stanford.edu} \\
  \and
  Hang Zhao $\P$ \\ 
  Tsinghua University \\ 
  \texttt{hangzhao@mail.tsinghua.edu.cn} \\
}


\maketitle

\begin{abstract}
   \input{chapters/abstract}
\end{abstract}

\section{Introduction}
\input{chapters/intro}

\input{chapters/related}

\input{chapters/methods/method}

\input{chapters/experiments/exp}
\input{chapters/conclusion}

{\small
\bibliographystyle{ieee_fullname}
\bibliography{egbib}
}

\end{document}

%% file: chapters/abstract.tex

Accurate and consistent 3D tracking from multiple cameras is a key component in a vision-based autonomous driving system. It involves modeling 3D dynamic objects in complex scenes across multiple cameras. This problem is inherently challenging due to depth estimation, visual occlusions, appearance ambiguity, etc. Moreover, objects are not consistently associated across time and cameras. To address that, we propose an end-to-end \textbf{MU}lti-camera \textbf{TR}acking framework called MUTR3D. In contrast to prior works, MUTR3D does not explicitly rely on the spatial and appearance similarity of objects. Instead, our method introduces \textit{3D track query} to model spatial and appearance coherent track for each object that appears in multiple cameras and multiple frames. We use camera transformations to link 3D trackers with their observations in 2D images. Each tracker is further refined according to the features that are obtained from camera images. MUTR3D uses a set-to-set loss to measure the difference between the predicted tracking results and the ground truths. Therefore, it does not require any post-processing such as non-maximum suppression and/or bounding box association. MUTR3D outperforms state-of-the-art methods by 5.3 AMOTA on the nuScenes dataset. Code is available at: \url{https://github.com/a1600012888/MUTR3D}. 


%% file: chapters/intro.tex

\begin{figure*}[t]
    \centering
    \includegraphics[width=1.95\columnwidth]{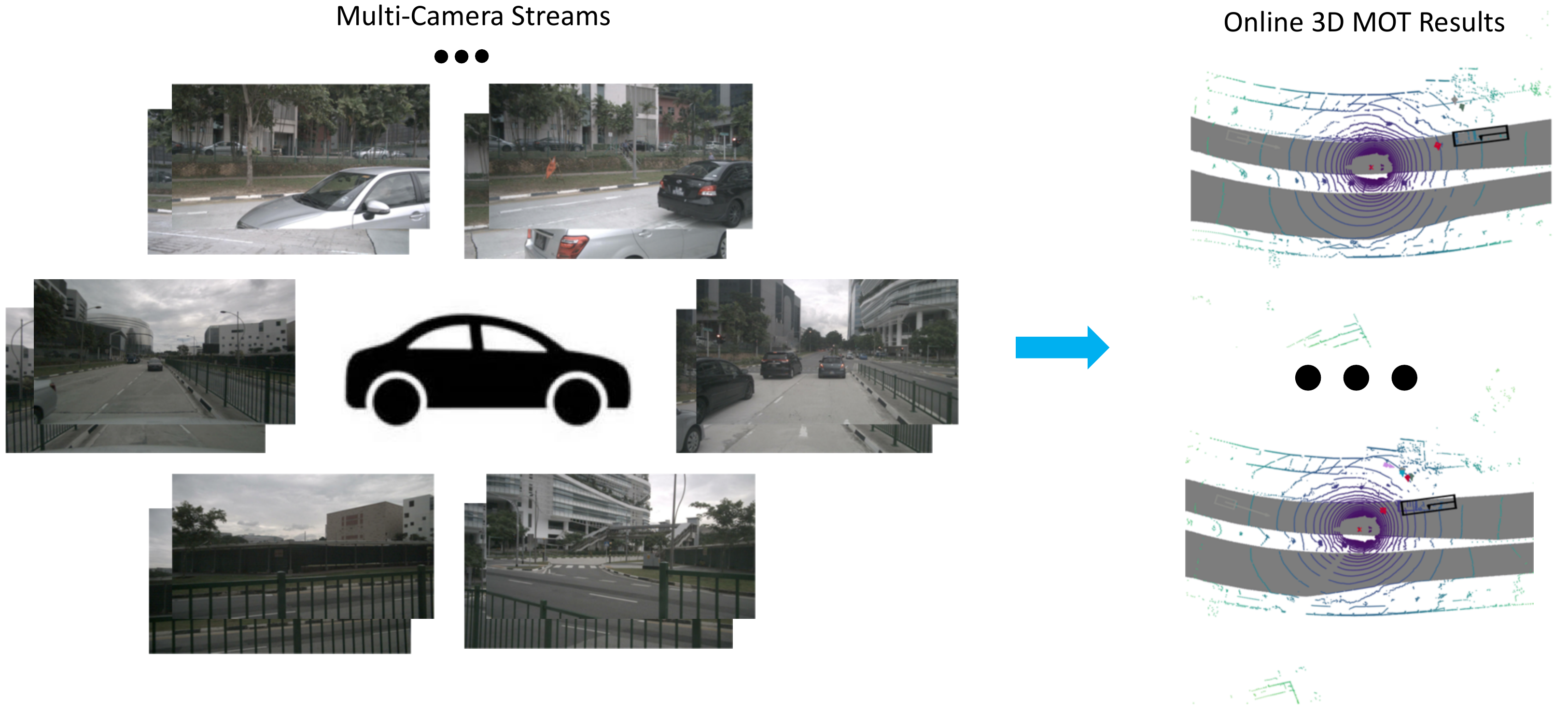}
    \caption{We propose an end-to-end Multi-camera 3D tracking framework, named MUTR3D. Our algorithm works with arbrtary camera rigs with known parameters. It handles multi-camera 3D detection, and cross-camera,  cross-frame objects association end-to-end fashion.  }
    \label{fig:teaser}
\end{figure*}

3D tracking is crucial in various perception systems, such as autonomous driving, robotics, and virtual reality. 
In its most basic incarnation, 3D tracking involves predicting per-frame objects and finding the correspondences between them temporally. Given per-frame object detection results, this problem boils down to associating objects across frames in a coherent fashion according to object similarity. On the other hand, tracking improves detection stability and enforces consistency of detection predictions across frames. 
However, this induces a complicated iterative optimization problem. 

More challenges arise when detailing multi-camera cases. First, accurate 3D detection is necessary for accurate tracking. However, camera-based 3D object detection remains an unsolved problem. Second, vision trackers are fragile regarding occlusion and appearance ambiguity in complex scenes. For example, a person of interest may walk behind a car and re-appear after a couple of seconds in a different pose. Third, trackers often lose objects moving across camera view boundaries. Therefore, beyond temporal association, we need to perform cross-camera association when objects span or cross different cameras to make spatially consistent predictions. These challenges hamper the practical use of 3D vision trackers. 

There are only a handful of works on vision-based 3D object tracking. Classical Kalman filtering-based methods~\cite{am3dmot} take detection results from any detectors as input and further make object state estimation and associations across time. More recent learning-based methods also follow a detect-to-track paradigm, where they first perform object proposals for each frame and then associate them in the feature space with a deep neural network~\cite{wojke2017simple,deft,qd3dt}.  

In this work, we propose MUTR3D, an online multi-camera 3D multi-object tracking framework that associate objects into 3D tracks using spatial and appearance similarities in an end-to-end manner. More concretely, we introduce \textit{3D track query}, which directly models the 3D states and appearance features of an object track over time and across cameras. At each frame, a 3D track query sample features from all visible cameras, and learn to create/track/end a track. In contrast to previous works, MUTR3D performs detection and tracking simultaneously in a unified and end-to-end framework. Objects decoded from the same queries across frames are inherently associated.

In summary, our contributions are three-fold:
\begin{itemize}
    \item To the best of our knowledge, MUTR3D is the first fully end-to-end multi-camera 3D tracking framework. Unlike existing detect-to-track methods that use explicit tracking heuristics, our method implicitly models the position and appearance variances of object tracks. Furthermore, we simplify the 3D tracking pipeline by eliminating commonly used post-processing steps such as non-maximum suppression, bounding box association, and object re-identification (Re-ID). 
    \item We introduce a \textit{3D track query} which models the 3D states of the entire track of an object. 3D track query samples feature from all visible cameras and update the track frame-by-frame end-to-end.
    \item Our end-to-end 3D tracking method achieves state-of-the-art performance on NuScenes vision-only 3D tracking dataset with 27.0\% AMOTA. More specifically,  MUTR3D performs much better than previous SOTA methods in the multi-camera setting with 12\% less ID switch. 
    \item We propose two metrics to evaluate motion models in the current 3D tracker: Average Tracking Velocity Error (ATVE) and Tracking Velocity Error (TVE).  They measure the error in the estimated motion of tracked objects. 
\end{itemize}

%% file: chapters/related.tex
\section{Related Work}

\subsection{3D MOT in Autonomous driving}
For autonomous cars, it is critical to track surrounding objects while estimating their position, orientation, size, and velocity. Due to recent advances on 3D detection \cite{pointpillar,voxelnet,second,centerpoint,monodis}, modern 3D MOT follows tracking-by-detection paradigm. These methods detect objects in the current frame and then associate them with previous tracklets. Weng \etal \cite{am3dmot} benchmark a simple yet effective association methods. They predict the location of previous tracklets through Kalman filtering, then associate current detections using 3D IoU.
Beyond IoU, several works used L2 distance \cite{centerpoint} and generalized 3D IoU\cite{simpletrack} to associate 3D box with pure location cues.  
Many works use more advanced association by adding learned motion and appearance features \cite{fantrack,deft,prob3dmot} or using graph neural networks \cite{ogr3mot,gnn3dmot,mpnmot}. Several works study how to improve life cycle management \cite{cbmot,simpletrack} by utilizing cues from detection scores. QD3DT current SOTA (State-of-The-Art) camera-based tracking algorithms learn an appearance matching feature through dense contrastive learning  They use an LSTM-based motion model to learn motion features and predict current locations. Finally, it combines visual features, motion cues, and depth-ordering for the association. Though with strong RGB appearance cues, performance of camera-based 3D MOT \cite{mono3dmot,fantrack,centertrack,deft,qd3dt,permatrack} has been lagged behind compared to LiDAR-based. On the nuScenes 3D MOT challenge's public leaderboard, STOA camera-based methods achieve 21.7\% AMOTA while STOA LiDAR-based methods reach 67.9\% AMOTA.  The problem of tracking through multiple distinct viewpoints also draws attention \cite{tan2019multi}. 

\subsection{Camera-based 3D Detection}

3D object detection have seen great advances in recent years. A stream of algorithms build upon 2D detection framework \cite{3dssd,monodis,centernet,fcos3d}. To resolve the fundamental ambiguity of instance depth and scales,  categorical canonical shapes\cite{geometry3ddet,metricshape3ddet},  geometric relation graphs\cite{pgd3ddet} and pretrained monocular depth \cite{rethinkpl,dd3d} are used. 
Another stream of methods works with representations on 3D space or Birds-Eye-View. Pseudolidar \cite{pseudolidar,pseudolidarpp} use pre-trained monocular depth models to lift pixels to 3D point clouds, then perform 3D detection using a LiDAR-based detector. Lift-Splat-Shot \cite{lss} makes the lifting process fully differentiable and joint trains the lifting modules with downstream tasks. Later CaDDN\cite{categoricaldepth} and BEVDet\cite{bevdet} used similar represents for 3D detection. DETR3D\cite{detr3d} adopt an inverse projecting process and build query-based multi-camera 3D detectors.
Compared to working on perspective image planes directly, one major advantage of working in 3D space is the ease of adopting arbitrary camera rigs and fusing multiple sensor features. Currently, there is no clear advantages on performance\cite{dd3d}. More comparisons are still yet under-explored.

\subsection{Query based detection and tracking}

A dominant type of modern detection and tracking approach is to reduce the task of detection to pixel-wise regression and classifications\cite{fasterrcnn,yolo,focalloss,densebox,centernet,fcos}, then perform tracking by associating detection boxes. Recently, DETR\cite{detr} successfully used query-based set prediction to achieve state-of-the-art detection results. Later TrackFormer\cite{trackformer}, MOTR and TransTrack\cite{motr, transtrack} extends this idea to online 2D MOT.  Our work builds upon the framework of query-based tracking. We extend the framework to multi-camera 3D MOT with a motion model.

%% file: chapters/methods/method.tex
\section{Methods}

\begin{figure*}[t]
    \centering
    \includegraphics[width=2.0\columnwidth]{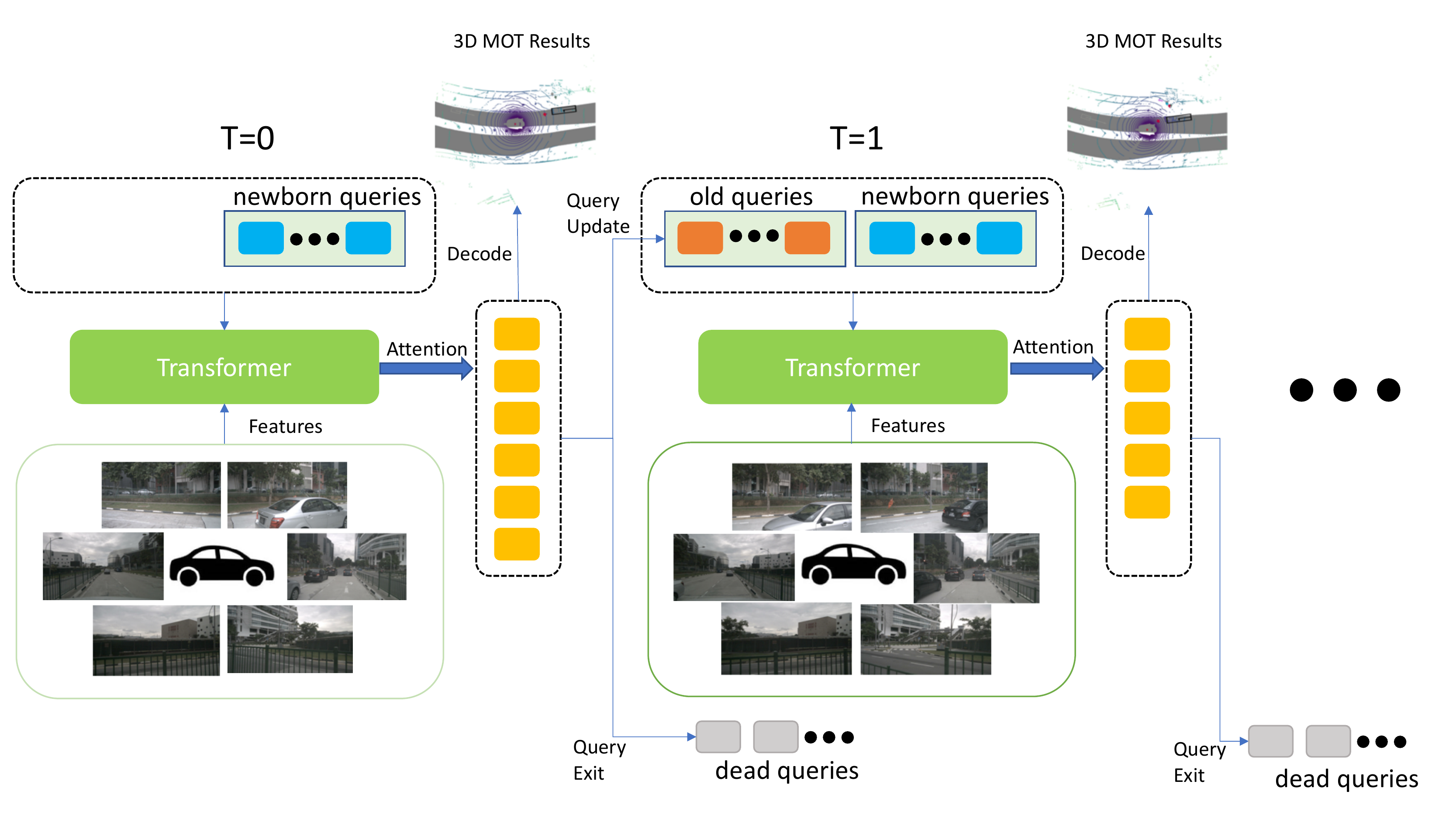}
    \caption{Pipeline overview of our online multi-camera tracker. All small colored squares in the black dashed box represent track queries. Blue boxes represent newborn queries, a fixed-set of learnable queries added to the set of track queries at the beginning of each frame. Orange boxes denote old queries, which are active queries from previous frames. Track queries attend with multi-camera features to decode object candidates in the current frame. Then we filter out inactive queries. We also update reference points of active queries to compensate for object motions and ego-motion. Finally, the updated queries went to the following frames to track the same objects.  }
    \label{fig:arch}
\end{figure*}




\subsection{Query based Object tracking}

We adopt query-based tracking for our algorithms. Query-based tracking is extended from query-based detection\cite{detr}, where detect queries, a fixed-size set of embedding, are used to represent 2D object candidates.   Track query extends the concept of the detect query to multi-frames, i.e., representing a whole tracklet across frames \cite{motr,trackformer,vistr}.
Specifically, we initialize a set of newborn queries at the beginning of each frame, then queries update themselves frame-by-frame in an auto-regressive way. A decoder head predicts one object candidate from each track query in each frame, and boxes decoded in different frames from the same track query are directly associated.   With proper query life cycle management, query-based tracking can perform joint detection and track in an online fashion. 


There are three key ingredients in our query-based multi-camera 3D tracker. (1) A query-based object tracking loss assigns different regression targets for two different types of queries, newborn queries, and old queries. (2) A multi-camera sparse attention uses 3D reference points to sample image features for each query.  
(3) A motion model estimates object dynamics and updates the query's reference point across frames.  We illustrate the flow of our trackers in Figure~\ref{fig:arch}.

\subsection{End-to-end object tracking loss}
\label{sec:loss}


We first explain the concept of \textit{label assignment} in the context of query-based tracking. Our algorithms maintain a changing set of track queries across frames. At the current frame, we decode one object candidate from each query. Ideally, The decoded object candidates from the same query should represent the same object across frames, thus forming a whole tracklet. 
To train the query-based tracker, we need to assign one target ground truth object for each query in each frame, and the assigned ground-truth object acts as the regression target for the query. 
Specifically, label assignment is a mapping function between ground-truth objects and track queries. We typically pad the set of ground truth objects with $\noobject$ (no object) to the number of predicted object candidates to ensure the mapping is a one-to-one mapping. Suppose we have $N$ decoded object candidates $\{\hat{y}_{1}, \ldots \hat{y}_{N} \}$ in current frame, label assignment can be denoted as a mapping  $\pi \in \{1,2 \ldots, N \} \mapsto \{1,2 \ldots, N \} $. Then the training loss can be expressed as a sum of paired box loss: 


\begin{equation} \label{eq:loss}
\mathcal{ L} = \sum_{i=1}^{N} \mathcal{ L}_{\mathrm{box}} (y_{\pi(i)}, \hy_{i}),
\end{equation}
where $y_{\pi(i)}$ denotes the assigned target ground-truth object, and $\mathcal{ L}_{\mathrm{box}}$ could be any bounding box loss. 
There are two types of queries for each frame, and they have different label assignment strategies. \textit{Newborn queries} are a set of learned queries. They are input-agnostic and will be added to the set of queries at the beginning of each frame. Newborn quires are responsible for detecting newly appeared objects in the current frame. So we perform bipartite matching between object candidates from newborn queries with newly appeared ground truth objects as DETR\cite{detr}.
\textit{Old queries} are active queries from previous frames which successfully detected or tracked objects. Old queries are responsible for tracking previously appeared objects in the current frame. The assignment for old queries is fixed after the first time it successfully detected a ground truth object. It is assigned to track the same object if they are in the current frame; otherwise, $\noobject$ (no object).

The 3D box loss $\mathcal{ L}_{\mathrm{box}}$ in equation~\ref{eq:loss} is defined as:

\begin{small}
\begin{equation}
    \mathcal{ L}_{\mathrm{box}}(y_{\pi(i)} , \hy_{j}) = \begin{cases}
    L_{\mathrm{cls}}(c_{\pi(i)} , \hat{c}_{j}) + \lambda L_{\mathrm{reg}}(b_{\pi(i)}, \hb_{j})  &  y_{\pi(i)} \neq \noobject \\
    L_{\mathrm{cls}}(c_{\pi(i)} , \hat{c}_{j}) & y_{\pi(i)} = \noobject \\
    \end{cases}
\end{equation}
\end{small}
We use $L_1$ loss for $ L_{\mathrm{reg}}$, and $L_{\mathrm{cls}}$ is the focal loss \cite{focalloss}, and the 3D object $y_{\pi(i)}$ is parameterized using the class label  $c_{\pi(i)}$, and bounding box parameters $b_{\pi(i)}$, details for the parameterization is in equation \ref{eq:box-center}.


\subsection{Multi-camera Track query decoding}


Our transformer decoder head takes track queries and attends them with multi-camera image features, and the extracted query featurees would be used to decode object candidates. Our decoder has two types of attention modules: self-attention between queries and cross attention between queries and image features. For memory efficiency, we adopt a reference-point based attention from DETR3D\cite{detr3d} for cross attentions. 
For notation in this section, only 3D coordinates or their 2D projections are in bold, e.g., 3D coordinates of reference points, $\bc_i$, estimated velocities $\bv_i$.  

\paragraph{\textbf{Query initialization. }} 
We assign a 3D reference point $\bc_{ i}$ to each query when it is initialized, i.e., when it is introduced as a newborn query at a certain frame.  The 3D reference point is decoded from its learnable embedding using a shared MLP (multi-layer perceptrons):

\begin{equation}\label{eq:query-point}
      \bc_{ i}=\Phi^{\mathrm{ref}}(\bq_{ i}),  
\end{equation}
where $\bq_{ i}$ denotes the learnable query embedding, the 3D reference points would be updated auto-regressively through layers of transformer decoders and across frames. It aims to approximate the 3D location of an object candidate.  

\vspace{-0.5em}
\paragraph{\textbf{Query feature extraction.}}

The cross attention works by projecting the reference point of each query to all the cameras and sampling point features. Suppose we have synchronized images from $M$ cameras for each frame. We extract pyramidal features for each image independently. We denote the set of pyramidal features as:   $\F_{1}, \F_{2}, \F_{3}, \F_{4}$.  
Each item 
$\F_{k}=\{F_{k1}, \ldots,F_{kM}\}, F_{ki} \in \R^{H\times W \times C}$
corresponds to a level of features of the $M$ images. We denote the provided camera projection matrices as $\T=\{T_1, \ldots, T_M\}, T_i \in \R^{3\times4}$. Specifically, the sampled point feature $f_{\bc_i}$ is : 

\begin{equation}\label{eq:img-weighted-sum}
    \begin{split}
        \bc_{ m i} & = T_m (\bc_{ i} \oplus 1),  w_{i}  = \mathrm{MLP}(\bq_{i}) , \\
        f_{\bc_i} & = \sum_k^{4} \sum_m^{M} F_{km} (\bc_{mi})\cdot  \sigma (w_{k m i}), 
    \end{split} 
\end{equation}
where $\bc_{ m i}$ denotes the projected 2D coordinates on the image plane of camera $m$,  $F_{km} (\bc_{mi})$ represents bilinear sampling from image features, and $\sigma()$ denotes sigmoid function, which is used to normalize the weighting factor. 

Then we use the extracted feature to update the query and its reference point 
\begin{equation}\label{eq:ref-update}
    \begin{split}
        \bq_{i} \leftarrow &  \bq_{i} + \mathrm{MLP} (f_{\bc_i} + \mathrm{PE}(\bq_{i} )), \\ 
        \bc_{i} \leftarrow &  \bc_{ i} + \mathrm{MLP} (f_{\bc_i}),
    \end{split} 
\end{equation}
where $\mathrm{PE}$ is learnable positional encoding, it is initialized with each query.  After layers of transformer decoder, we use the final query feature to decode object candidate in the current frame. 

\vspace{-0.5em}
\paragraph{\textbf{3D Object Parametrization.}} \label{subsec:object-param}
We use two small FFNs to decode 3D box parameters and categorical labels. We parameterize the 3D box by additional ten dimensional parameters: coordinates of the box center in ego frame, $x_{ i} \in \R^3$, size of the 3D box $s_{ i} = (w_{ i}, l_{ i}, h_{ i}) \in \R^3$, 2D velocity in ego frame  $v_{ i} = (v_{i}^x, v_{i}^y) \in \R^2 $ and orientation $(\sin{\theta_{ i}}, \cos{\theta_{ i}})$, where $\theta_{ i}$ is the yaw-angle in ego frame. The coordinates of the box center is predicted by adding a residual to the reference point:

\begin{equation}\label{eq:box-center}
    \bx_{ i} = \bc_{ i} + \mathrm{MLP}(\bq_{i}) .
\end{equation}

\subsection{Query Life Management}
To deal with disappearing objects in an online fashion, we need to remove inactive queries after each frame. We define the confidence score of each query as the classification score of their predicted box.  We use two threshold parameters $\tau_{\mathrm{new}}$ and $\tau_{\mathrm{old}}$ for box scores and a time length, $T$ to control the life management. 

During inference, for newborn queries in each frame, if the score is lower than $\tau_{\mathrm{new}}$, we remove it.  For old queries, if their scores have been lower than  $\tau_{\mathrm{old}}$  for successive $T$ frames, we remove it. We select  $\tau_{\mathrm{new}} = 0.4$, and $\tau_{\mathrm{old}} = 0.35$ and $T=5$ for nuScenes dataset after several trails.


During training, we view queries matched to $\noobject $ as inactive. For newborn queries in the current frame, if it is matched to $\noobject$, we remove it. For old queries, we remove it if it has been matched to $\noobject$ for successive $T$ times. Note that old queries that have been matched to $\noobject$ but have not been removed continue to update themselves through the transformer decoder.

\subsection{Query Update and Motion model}

After filtering out outdated (dead)  queries, we update track queries, both their features and 3D reference points. The purpose of updating the 3D reference point is to model object dynamics and compensate for ego-motion. There are two commonly used motion models in 3D tracking, Kalman Filter, e.g., \cite{am3dmot,simpletrack}, which uses observed position across frames to estimate unknown velocity, and predicted velocity from detectors, e.g., CenterTrack \cite{centertrack,centerpoint}. We use velocity predicted from queries, which updates through frames and can aggregate multi-frame features. We use a small FFN to predict ego frame velocity. The predicted velocity is supervised with ground truth. Denote the ego pose of current frame and next frame as  $R_t, R_{t+1} \in R^{3\times 3}, T_{t}, T_{t+1} \in R^{3}$.  Denote the time gap between these two frames as $\Delta t$. We update the reference point $\bc_{i}$ of the $i$-th query using the predicted box velocity $\bv_{ i} = (v_{i}^x, v_{i}^y, 0) \in \R^3 $ : 

\begin{equation} \label{eq:velo-update}
\bc_{i} \leftarrow R_{t+1}^{-1} ( R_{t} (\bc_{i} + \bv_{i} \times \Delta t ) + T_{t} - T_{t+1}). 
\end{equation}

To implicitly model multi-frame appearance variations, we update the track query using features from previous frames.  Following MOTR\cite{motr}, we maintain a fixed-size first-in-first-out queue for each of the active queries, named \textit{memory bank}.
After each frame, we apply an attention module for each query and its memory bank.  The track query acts as the query for the attention module, and the corresponding memory bank act as a set of keys and values.


%% file: chapters/experiments/exp.tex
\section{Experiments}

\subsection{Datasets}
We use nuScenes\cite{nuscenes} dataset for all of our experiments. It consists of 1000 real-world sequences, 700 sequences for training, 150 for validation, and 150 for the test. Each sequence has roughly 40 annotated keyframes. Keyframes are synchronized frames for each sensor with a sampling rate of 2 FPS. Each frame includes images from six cameras with a full 360-degree field of view. It provides 3D tracking annotations for 7 Object categories. 

\input{chapters/experiments/table-sota}

\subsection{Evaluation Metrics}
Average multi-object tracking accuracy (AMOTA) and average multi-object tracking precision (AMOTP) are the major metrics for nuScenes 3D tracking benchmark.  AMOTA and AMOTP are computed by integrating MOTA(multi-object tracking accuracy) and MOTP(multi-object tracking precision) values over all recalls: 
\begin{equation}\label{eq:amota}
    \mathrm{AMOTA} =   \frac{1}{L} \sum_{r \in \{\frac{1}{L}, \frac{2}{L}, \ldots, 1 \}} \mathrm{MOTA}_r,
\end{equation}

\begin{equation}
    \mathrm{MOTA}_r = \max{( 0, 1 - \frac{ \mathrm{FP}_r + \mathrm{FN}_r + \mathrm{IDS}_r - (1-r) \mathrm{GT}}{r \mathrm{GT}} )}, 
\end{equation}
where $\mathrm{FP}_r$, $\mathrm{FN}_r$ and $\mathrm{IDS}_r$ represents the number of false positives, false negatives, and identity switches computed at the corresponding recall $r$. $\mathrm{GT}$ is the number of ground truth bounding boxes. AMOTA can be formulated as:

\begin{equation}
    \mathrm{AMOTP} =  \frac{1}{L} \sum_{r \in \{\frac{1}{L}, \frac{2}{L}, \ldots, 1 \}} \frac{\sum_{i, t} d_{i, t}}{ \mathrm{TP}_{r}}, 
\end{equation}
where $d_{i, t}$ denotes the 2D birds-eye-view position error of matched track $i$ at time $t$, and $\mathrm{TP}_r$ indicates the number of matches computed at the corresponding recall $r$. 

We also report tracking metrics from CLEAR\cite{clear} and Li et al.\cite{li2009learning} such as MOTA, MOTP, IDS. The confidence threshold for these metrics is selected by independently picking the threshold with the highest MOTA for each category.

\subsection{Implementation Details}

\paragraph{Feature extractor}
Following prior works\cite{fcos3d} \cite{detr3d}
ResNet-101 with deformable convolutions\cite{dcn} and FPN\cite{fpn} are used for image feature extractors. For ablation study, we replace the ResNet-101 with ResNet-50 for memory efficiency. 

\paragraph{Training details} 
We use 3D detection pre-trained models from DETR3D\cite{detr3d}. Then we replace the head and train our tracker with three frames video clips for 72 epochs.  

\paragraph{Kalman filter baselines} Kalman filter-based methods have been state-of-the-art trackers on LiDAR-based 3D tracking across datsets\cite{simpletrack}. However, camera-based SOTA methods typically use learned appearance and motion features for matching. To further understand the field of camera-based 3D MOT, we provide two Kalman filter baselines with DETR3D\cite{detr3d} detector. 
(1) A basic version with no advanced design. The basic version improves over the public implementation of AB3DMOT \cite{am3dmot}. To handle the failure of IoU(Intersection over Union) during association with low frame rate data, we enlarge the prediction boxes by 20\% when computing 3D IoU. 
(2) We also provide an advanced version of Kalman filter baselines from SimpleTrack \cite{simpletrack}, which used 3D generalized IoU and two-stage associations. SimpleTrack obtained SOTA result on LiDAR-based MOT. 



\input{chapters/experiments/table-kf}

\subsection{Compare with State-of-the-art}
We compare our method with SOTA methods in Table~\ref{tab:sota}. We outperform current SOTA methods for the camera-based tracker by a large margin. The gain in AMOTA from the current SOTA method QD3DT\cite{qd3dt} is over 5.2 points on the validation set and 5.3 points on the test set. Our tracker operates in an end-to-end fashion, with no NMS and no association stages as in QD3DT\cite{qd3dt}.

We put the comparisons of two of our Kalman filter baselines in Table~\ref{tab:kf}. We outperform the basic version of the Kalman filter. However, when compared with more tailored baselines from SimpleTrack\cite{simpletrack}, we only have slight gains on metrics like AMOTA, MOTA, MOTP.

\input{chapters/experiments/table-velocity}

\subsection{Evaluating Motion Models}

The motion model provides one of the primary cues for 3D Multi-object Tracking. The motion model aims to describe the moving patterns of tracklets. To evaluate the motion models of different tracking algorithms, we develop two metrics, Average Tracking Velocity Error (ATVE) and Tracking Velocity Error (TVE), following the idea of AMOTP and MOTP. ATVE can be computed as: 
\begin{equation} \label{eq:atve}
    \mathrm{ATVE} =  \frac{1}{L} \sum_{r \in \{\frac{1}{L}, \frac{2}{L}, \ldots, 1 \}} \frac{\sum_{i, t} || v_i - v_t ||_2}{ \mathrm{TP}_{r}}, 
\end{equation}
where we traverse over all pairs of matched tracking predictions and ground truth and compute the $ L_2$  error between the predicted velocity $v_i$ and the ground-truth velocity $v_t$. Average Tracking Velocity Error is computed by averaging over all recalls $r$, and $\mathrm{TP}_{r}$ represents the number of matches in corresponding recall $r$. Like MOTP,  Tracking Velocity Error is the average velocity error computed at the recall with the highest MOTA. We evaluate the evaluation of motion models in Table~\ref{tab:velocity}. Compared to the previous state-of-the-art camera tracker QD3DT\cite{qd3dt}, our velocity is more accurate. Compared to Kalman filtering-based motion models, our algorithm achieves better Tracking velocity Error.

\input{chapters/experiments/table-ablation-motion}
\subsection{Ablation study}
We study two factors in the ablation study. First, we study the effect of dropping our motion model, i.e., do not update the 3D reference points at the end of each frame. We show the ablation results in Table~\ref{tab:motion}. Removing our motion model degrades the performance in all metrics. 

Second, we study the effect of the number of training frames. Our methods track objects in an auto-regressive way, and no teacher-forcing is applied. During training, gradients computed in latter frames will still propagate to compute graphs in previous frames. In the ablation study, we perform all the experiments using ResNet-50 backbones. We report the performance of training with 3,4,5 frames in Table~\ref{tab:training_frame}. Results showed increasing the number of training frames gradually improves the performance. 

\input{chapters/experiments/table-ablation-train-longer}
\vspace{-0.4em}

\input{chapters/experiments/figure}

\subsection{Qualitative results}

We provide visualizations of our tracking algorithms in both BEV and camera views for an 8 seconds clip in Figure~\ref{fig:quantitative}. Near-filed objects on the left/right side of the car are usually truncated by several cameras, which is a substantial challenge for multi-camera 3D tracking. See, the gray and black cars are truncated in the Front-Left camera and Back-Left camera(3-rd/7-th and 4-th/8-th row),  and our algorithm handles them correctly. 

%% file: chapters/experiments/table-sota.tex
\begin{table*}[t]
\caption{\textbf{Comparison with state-of-the-art methods on nuScenes dataset.} For public camera-based 3D tracking, our algorithm achieves state-of-the-art results, outperforming QD3DT\cite{qd3dt} by 0.052 in AMOTA on validation set and 0.053 on test split.   }
\label{tab:sota}
\centering
\scalebox{.9}{
\begin{tabular}{l| c | c c c c c | c}
\specialrule{1pt}{0pt}{1pt}

 & Modality & AMOTA $\uparrow$ & AMOTP $\downarrow$ & RECALL $\uparrow$ & MOTA $\uparrow$ & IDS $\downarrow$ & \#params \\

\midrule
Validation Split \\
\midrule

CenterPoint\cite{centerpoint}  & LiDAR & 0.665  & 0.567 & 69.9\% & 0.562 & 562 & 9M \\ 
SimpleTrack\cite{simpletrack}  & LiDAR & 0.687  & 0.573 & 72.5\% & 0.592 & 519 & 9M \\ 

\hline

DEFT \cite{deft} & Camera & 0.201 & N/A & N/A &  0.171 & N/A  & 22M \\

QD3DT\cite{qd3dt} & Camera & 0.242 & 1.518 & 39.9\% &  0.218 & 5646 & 91M\\
 
Ours & Camera & \textbf{0.294} & \textbf{1.498} & \textbf{42.7\%}  & \textbf{0.267}  & \textbf{3822} & 56M\\

\midrule
Test Split \\
\midrule
CenterTrack \cite{centertrack} & Camera & 0.046 & 1.543 & 23.3\% & 0.043 & \textbf{3807} & 20M \\

DEFT \cite{deft} & Camera & 0.177 & 1.564 & 33.8\% & 0.156 &  6901 & 22M \\

QD3DT\cite{qd3dt} & Camera & 0.217 & 1.550 & 37.5\% &  0.198 & 6856 & 91M\\
 
Ours & Camera & \textbf{0.270} & 1.494 & \textbf{41.1\%}  & \textbf{0.245}  & 6018 & 56M\\

\bottomrule
\end{tabular}
}
\end{table*}

%% file: chapters/experiments/table-kf.tex
\begin{table*}[ht]
\caption{\textbf{Comparison with Kalman Filter based methods on nuScenes validation split.} We construct two kalman filter baselines using our pretrained detector DETR3D\cite{detr3d}. We compare them with out tracker. }
\label{tab:kf}
\centering
\scalebox{.9}{
\begin{tabular}{l| c c c c c c}
\specialrule{1pt}{0pt}{1pt}

 & AMOTA $\uparrow$ & AMOTP $\downarrow$ & RECALL $\uparrow$ & MOTA $\uparrow$ & MOTP $\downarrow$ & IDS $\downarrow$ \\
\hline
DETR3D\cite{detr3d} + KF & 0.263  & 1.569  & 39.7\%  & 0.260 & 0.952 &  4698 \\

DETR3D + SimpleTrack\cite{simpletrack}  & 0.293 & \textbf{1.307} & 41.8\% & 0.263 & 0.84 & \textbf{1695} \\
Ours & \textbf{0.294} & 1.498 & \textbf{42.7\%}  & \textbf{0.267}  & \textbf{0.799} & 3822 \\
\bottomrule
\end{tabular}
}
\end{table*}

%% file: chapters/experiments/table-velocity.tex
\begin{table}[ht]
\caption{\textbf{Evaluate velocity estimation.} We report ATVE (Average Tracking Velocity error) and TVE(Tracking Velocity error) for on nuScenes validation split. Compared with kalman filter based motion models, our methods obtain better TVE. }
\label{tab:velocity}
\centering
\scalebox{.9}{
\begin{tabular}{l| c | c  c }
\specialrule{1pt}{0pt}{1pt}

 & Modality & ATVE $\downarrow$ & TVE $\downarrow$\\


\hline

CenterPoint\cite{centerpoint}  & LiDAR & 0.572  & 0.298 \\ 

\hline

QD3DT\cite{qd3dt} & Camera & 1.876 & 1.373 \\
DETR3D + SimpleTrack & Camera &  \textbf{1.344}   & 0.836 \\
Ours & Camera & 1.548 & \textbf{0.768} \\
\bottomrule
\end{tabular}
}
\end{table}

%% file: chapters/experiments/table-ablation-motion.tex
\begin{table}[h]
\caption{\textbf{Ablation on  motion models.} When removing motion models, the performance of our algorithm drops in all metrics.}
\label{tab:motion}
\centering
\scalebox{.9}{
\begin{tabular}{l| c c c c c}
\specialrule{1pt}{0pt}{1pt}

 &  AMOTA  & AMOTP & RECALL & MOTA & IDS  \\


\hline

w/o Motion & 0.215 & 1.598 & 35.8\% & 0.198 & 4100   \\
w/ Motion & \textbf{0.234} &\textbf{ 1.585} & \textbf{38.7\%} & \textbf{0.22} & \textbf{3775}  \\
\bottomrule
\end{tabular}
}
\end{table}

%% file: chapters/experiments/table-ablation-train-longer.tex
\begin{table}[h]
\caption{\textbf{Ablation on the number of training frames.} Training our models with longer video clips is beneficial. }
\label{tab:training_frame}
\centering
\scalebox{.9}{
\begin{tabular}{l| c c c c c}
\specialrule{1pt}{0pt}{1pt}


\#frames &  AMOTA & AMOTP & RECALL & IDS & ATVE  \\

\hline

3 & 0.234 & 1.585 & 38.7\%   & \textbf{3775}  & 1.606 \\
4 & 0.242 & 1.580 & 39.7\%  & 4623  & 1.545 \\
5 & \textbf{0.251} & \textbf{1.573} & \textbf{39.9\%}   & 3873  & \textbf{1.565} \\
\bottomrule
\end{tabular}
}
\end{table}

%% file: chapters/experiments/figure.tex
\begin{figure*}[ht]
    \centering
    \vspace{-1em}
    \scalebox{.9}{
    \begin{tabular}{c c c c}
    \includegraphics[width=0.47\columnwidth]{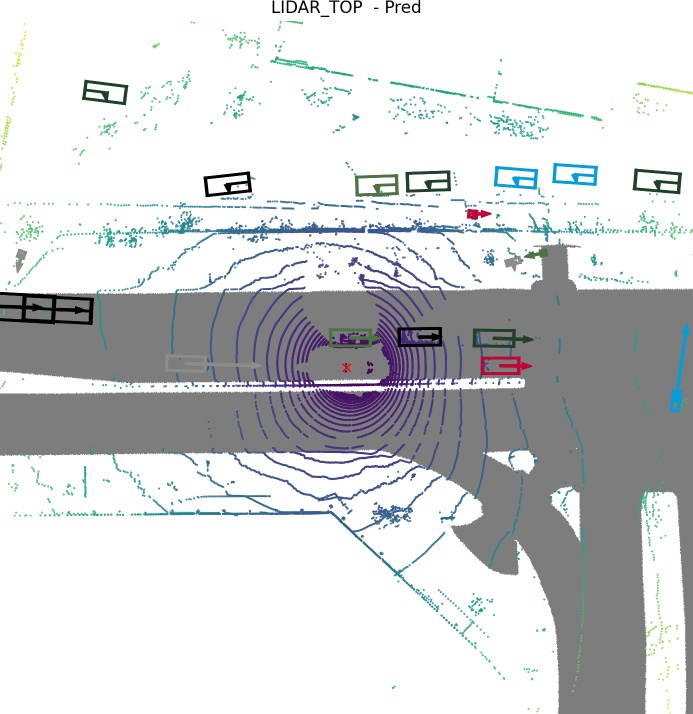}  & 
    \includegraphics[width=0.47\columnwidth]{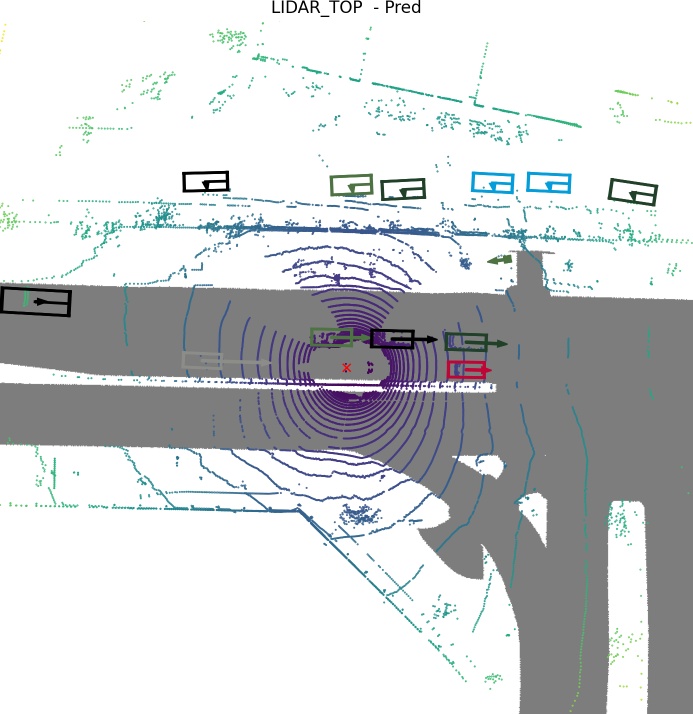}  &  \includegraphics[width=0.47\columnwidth]{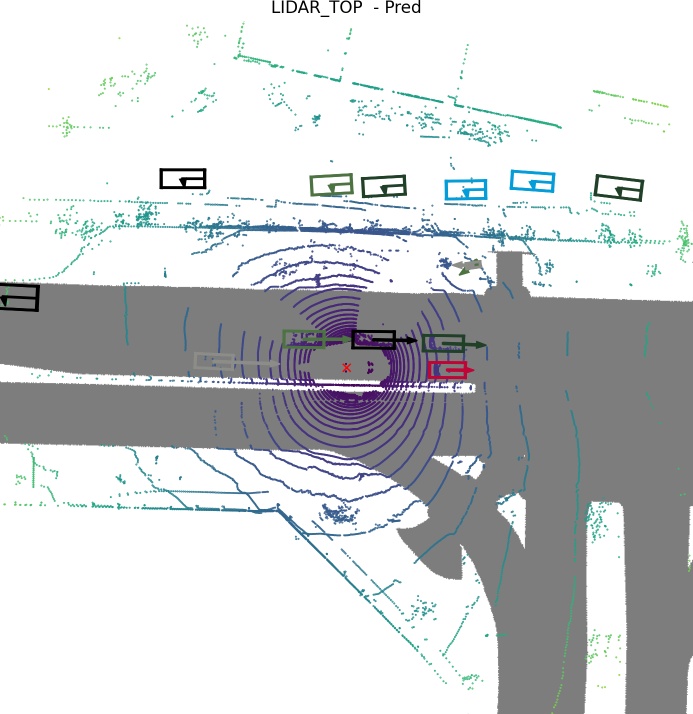} & 
    \includegraphics[width=0.47\columnwidth]{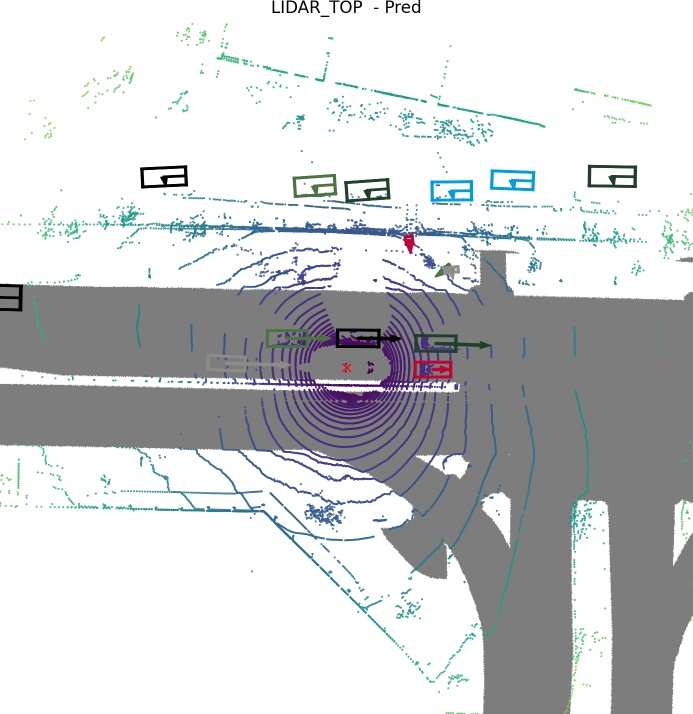} \\
         
         \includegraphics[width=0.47\columnwidth]{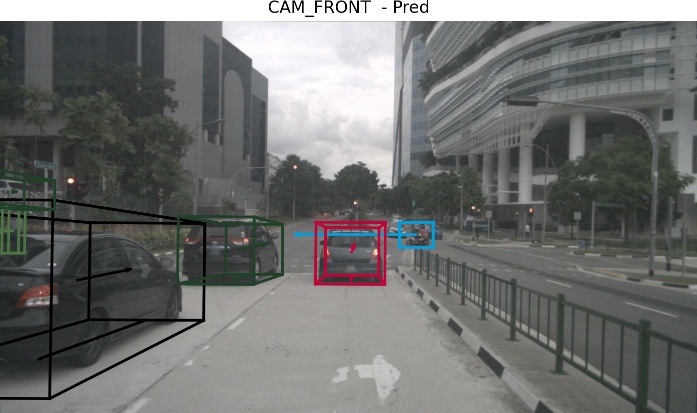}  & 
         \includegraphics[width=0.47\columnwidth]{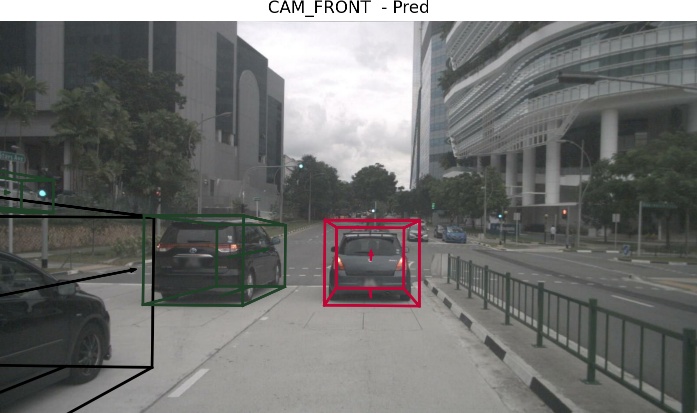}  &  \includegraphics[width=0.47\columnwidth]{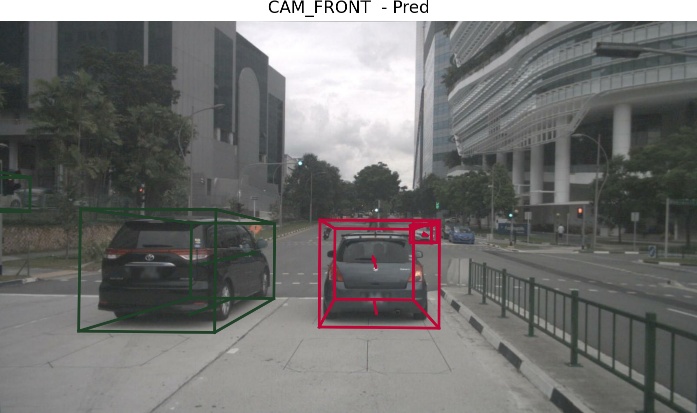} & 
         \includegraphics[width=0.47\columnwidth]{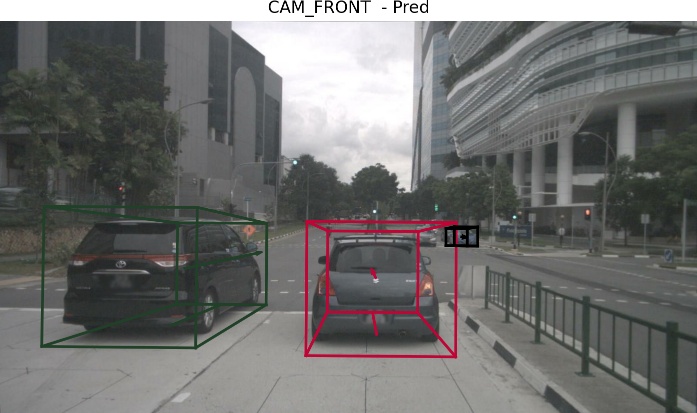}  
         \\
         
         \includegraphics[width=0.47\columnwidth]{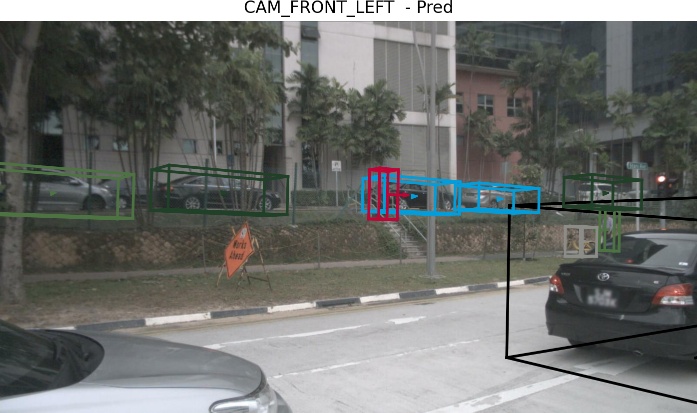} &
         \includegraphics[width=0.47\columnwidth]{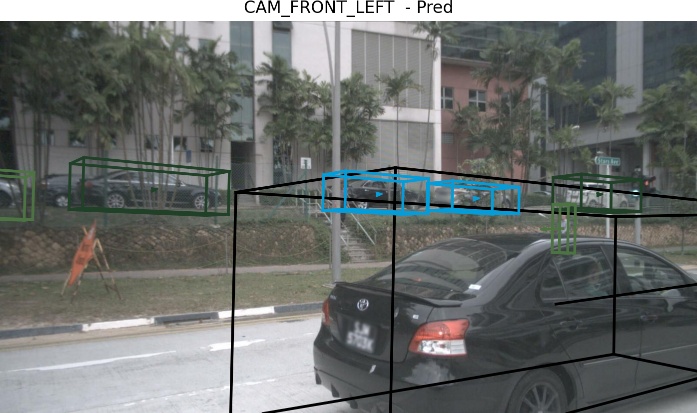} &  \includegraphics[width=0.47\columnwidth]{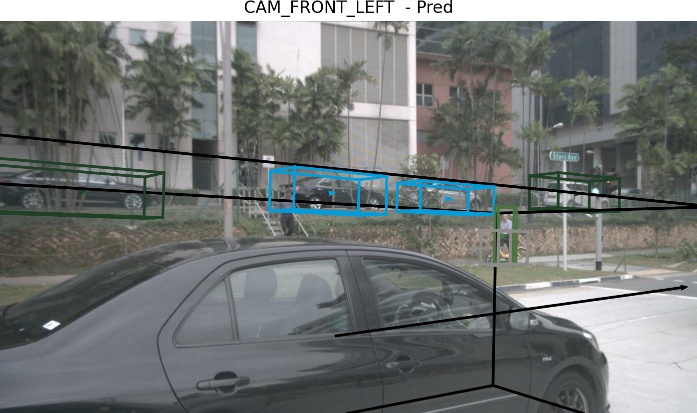} & 
         \includegraphics[width=0.47\columnwidth]{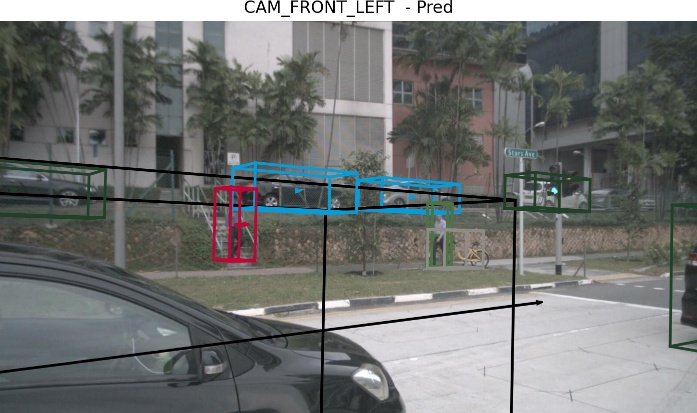} 
         \\
         
         \includegraphics[width=0.47\columnwidth]{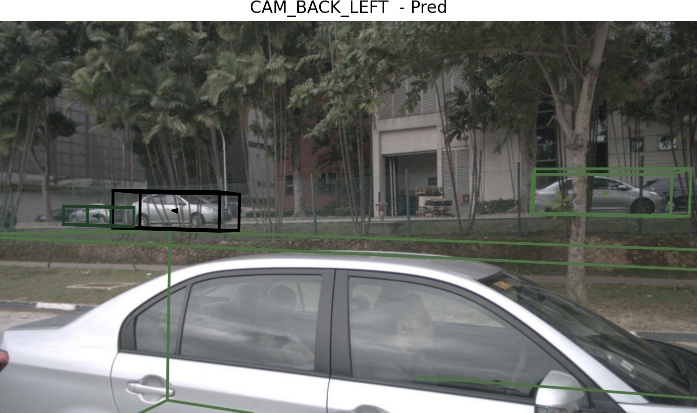}
         & \includegraphics[width=0.47\columnwidth]{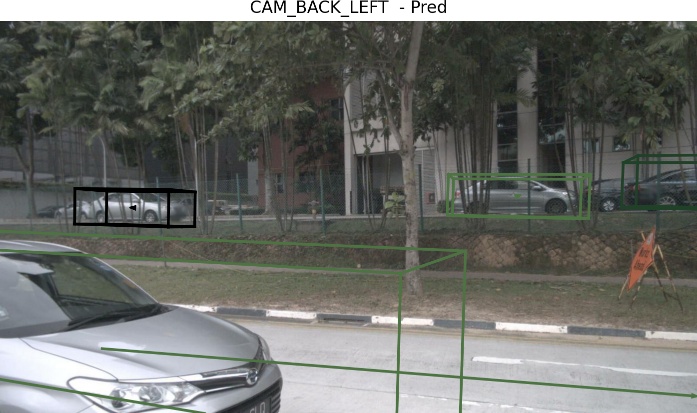} & \includegraphics[width=0.47\columnwidth]{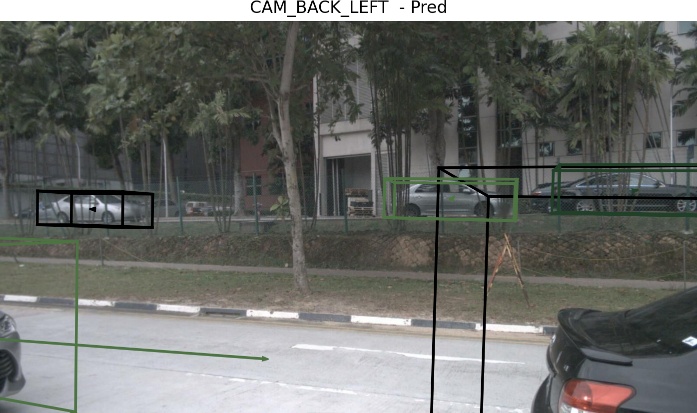} &
         \includegraphics[width=0.47\columnwidth]{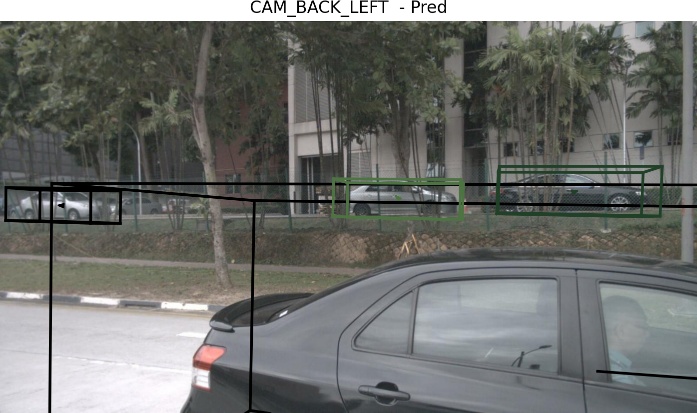}  
         \\
    \midrule
    
    \includegraphics[width=0.47\columnwidth]{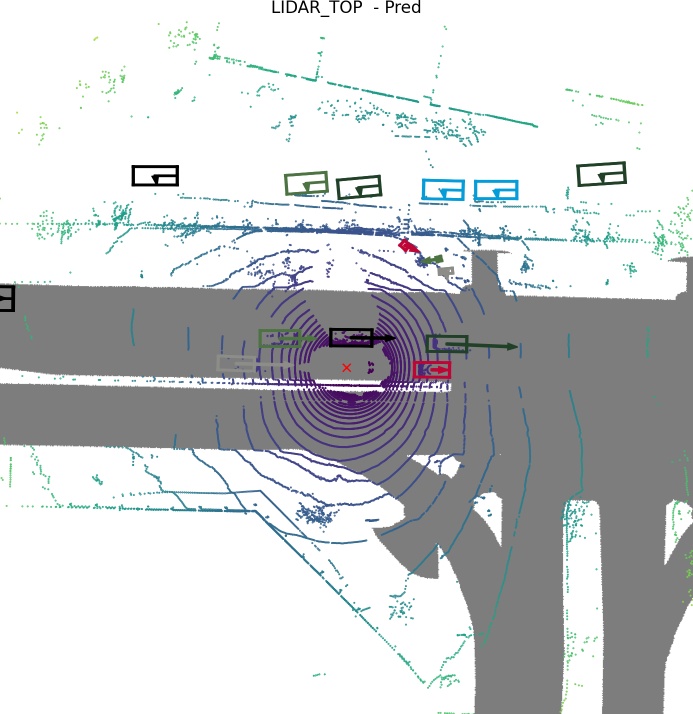} &
    \includegraphics[width=0.47\columnwidth]{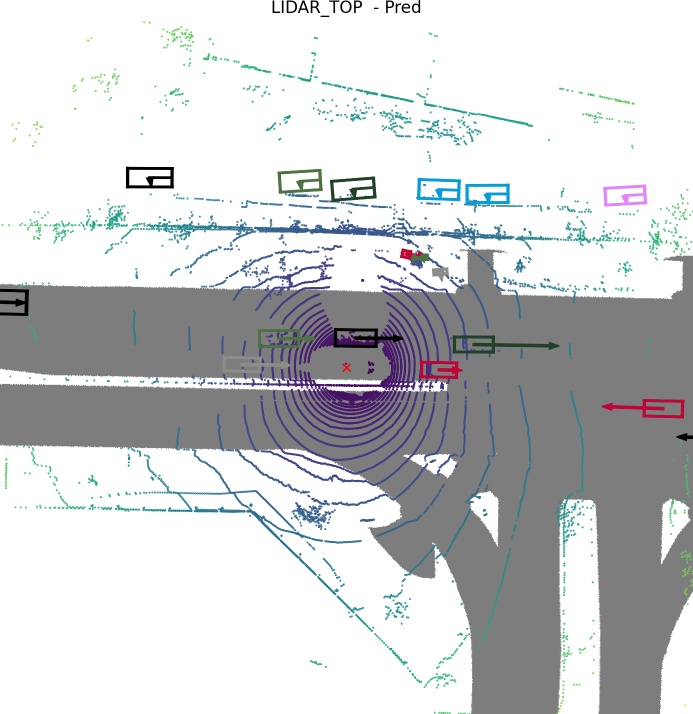}  &  \includegraphics[width=0.47\columnwidth]{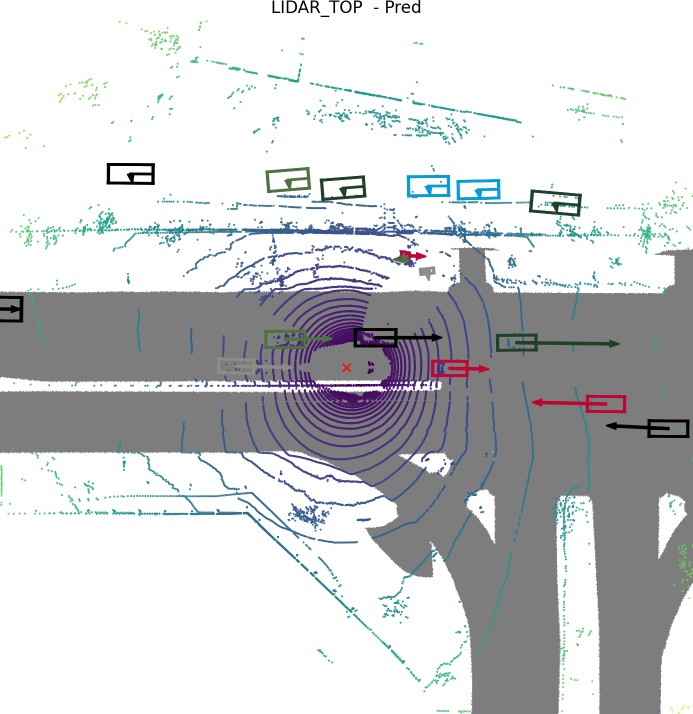} & 
    \includegraphics[width=0.47\columnwidth]{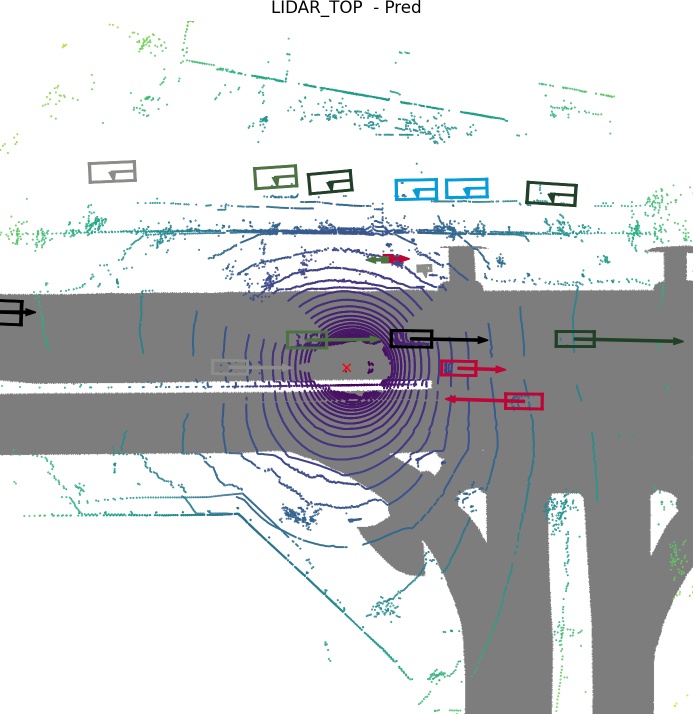} 
    \\
         
         \includegraphics[width=0.47\columnwidth]{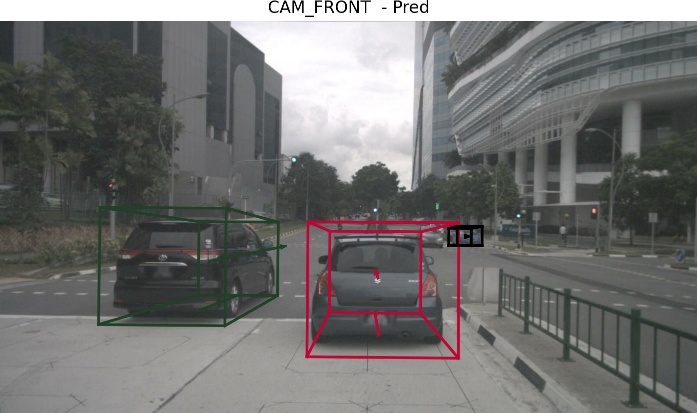} &
         \includegraphics[width=0.47\columnwidth]{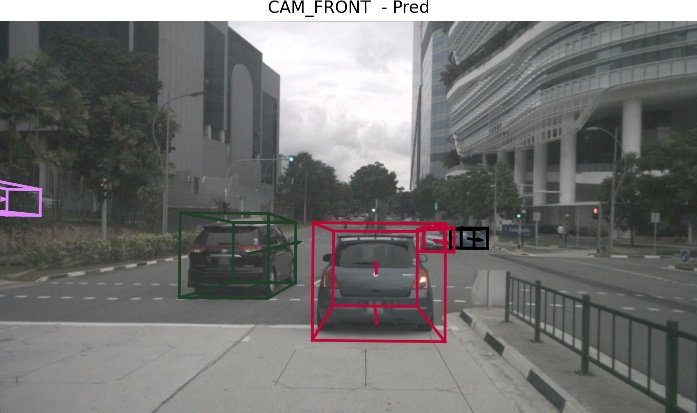}  &  \includegraphics[width=0.47\columnwidth]{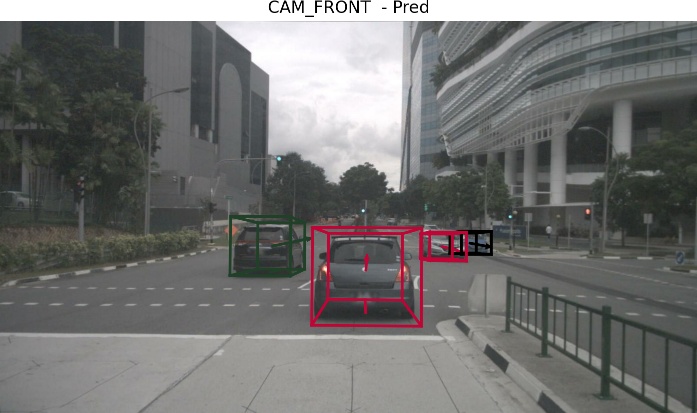} & 
         \includegraphics[width=0.47\columnwidth]{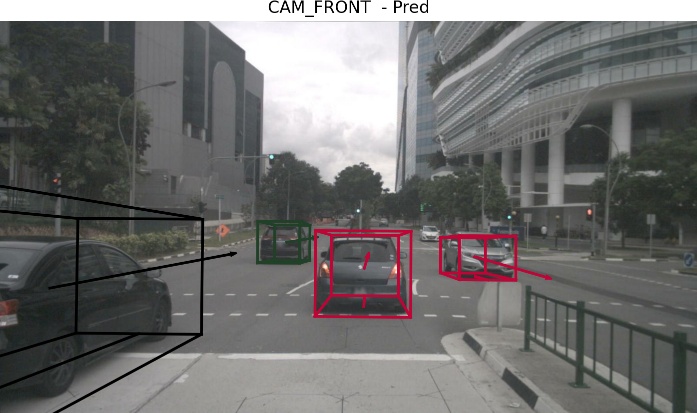}  
          \\
         
         \includegraphics[width=0.47\columnwidth]{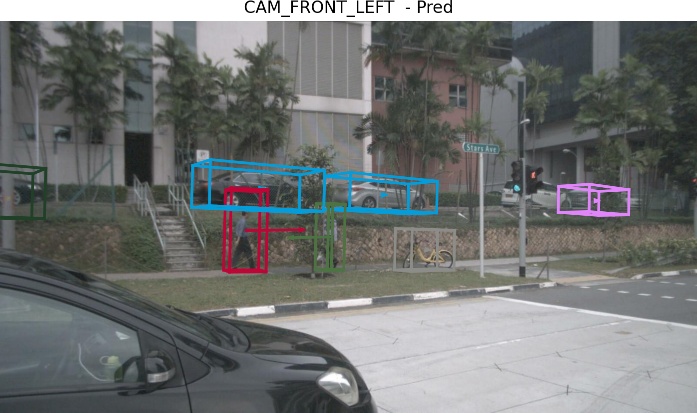} &
         \includegraphics[width=0.47\columnwidth]{figures/results/cam_front_left/34.jpg}  &  \includegraphics[width=0.47\columnwidth]{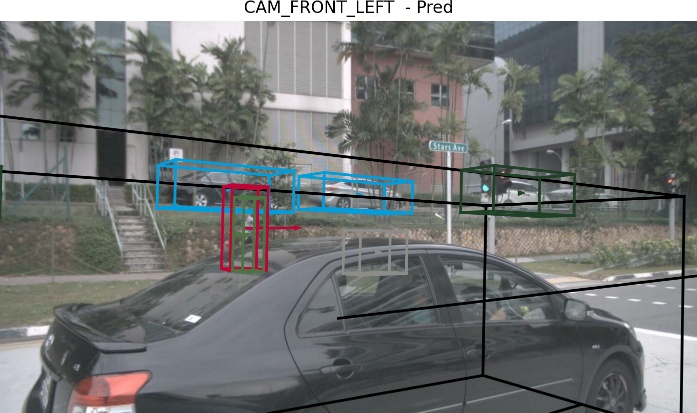} & 
         \includegraphics[width=0.47\columnwidth]{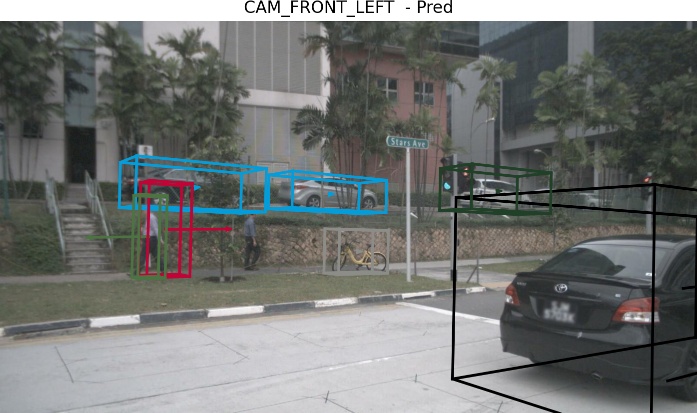}
          \\
         
         \includegraphics[width=0.47\columnwidth]{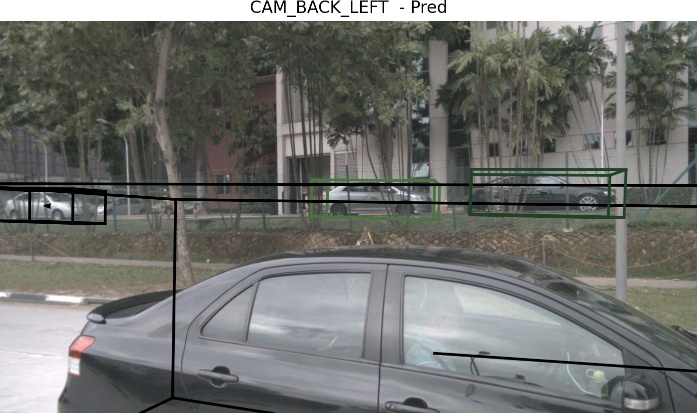} &
         \includegraphics[width=0.47\columnwidth]{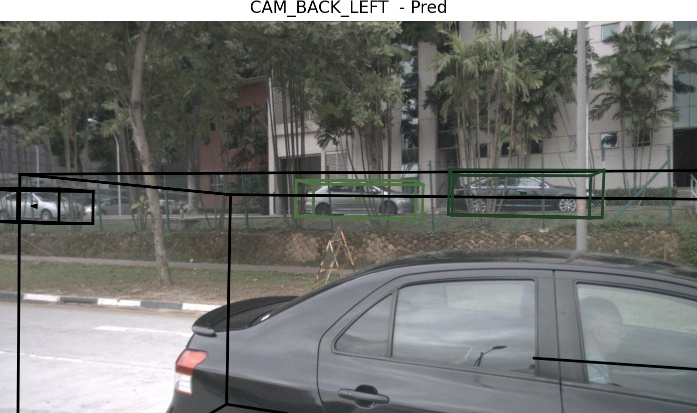}  &  \includegraphics[width=0.47\columnwidth]{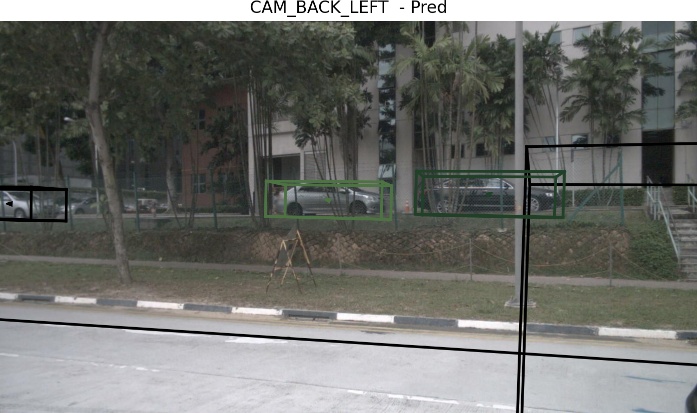} &
         \includegraphics[width=0.47\columnwidth]{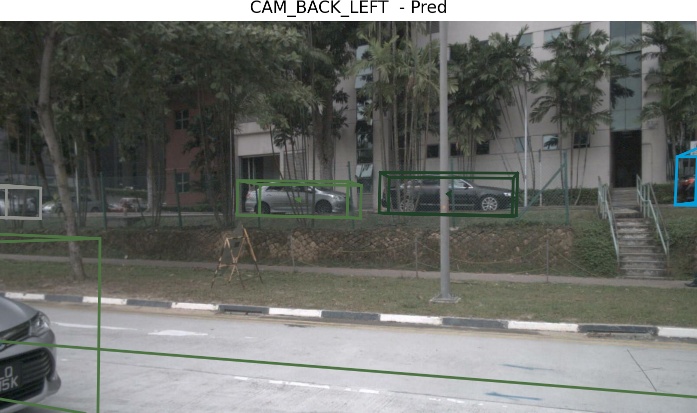}  
          \\
    
    \end{tabular}
    }
    \vspace{-0.5 em}
    \caption{Visualization on 8 consecutive frames with FPS as 1. We plot the results for 4 views, from top to bottom: Birds-Eye-View, Front camera, Front-left camera, and Back-left camera. Objects with the same identity are painted with the same color. We plot the estimated velocity using arrows, and longer arrows represent larger velocity. The example we showed contains multiple frames with truncated objects across cameras. Our algorithm are designed to fuse multi-camera features automatically, and handle the truncation correctly.   }
    \label{fig:quantitative}
\end{figure*}

%% file: chapters/conclusion.tex
\section{Conclusion}

We design an end-to-end multi-camera 3D MOT framework. Our framework can perform 3D detection, compensate for ego-motion and object motions, and perform cross-camera and cross-frame object association end-to-end. In the nuScenes test dataset, our tracker outperforms the current state-of-the-art camera-based 3D tracker QD3DT\cite{qd3dt} by 5.3 AMOTA and 4.7 MOTA. We also study the quality of the motion models in current 3D trackers by evaluating two new metrics: Average Tracking Velocity Error (ATVE) and Tracking Velocity Error (TVE). Compared to hand-designed associating methods, we believe our end-to-end learnable tracker can enjoy the abundant amount of data in autonomous driving fields in the future.